# Toward Personalized XAI: A Case Study in Intelligent Tutoring Systems

Cristina Conati[*a], Oswald Barral[a,] Vanessa Putnam[a], Lea Rieger[b],

[a]The University of British Columbia, Canada
[b]Augsburg University, Germany

**Abstract**

Our research is a step toward ascertaining the need for personalization, in XAI, and we do so in the context of investigating the value of explanations of AI-driven hints and feedback are useful in Intelligent Tutoring Systems (ITS). We added an explanation functionality for the adaptive hints provided by the Adaptive CSP (ACSP) applet, an interactive simulation that helps students learn an algorithm for constraint satisfaction problems by providing AI-driven hints adapted to their predicted level of learning. We present the design of the explanation functionality and the results of a controlled study to evaluate its impact on students' learning and perception of the ACPS hints. The study includes an analysis of how these outcomes are modulated by several user characteristics such as personality traits and cognitive abilities, to asses if explanations should be personalized to these characteristics. Our results indicate that providing explanations increase students' trust in the ACPS hints, perceived usefulness of the hints, and intention to use them again. In addition, we show that students' access of the explanation and learning gains are modulated by user characteristics, providing insights toward designing personalized Explainable AI (XAI) for ITS.

[*] corresponding author: conati@cs.ubc.ca

# 1. Introduction

Existing research on Explainable AI (XAI) suggests that having AI systems explain their inner workings to their end users[1] can help foster transparency, interpretability, and trust (e.g., [3]–[6]). However, there are also results suggesting that such explanations are not always wanted by or beneficial for all users (e.g., [7]–[9]). Our long-term goal is understanding when having AI systems provide explanations to justify their behavior is useful, and how this may depend on factors such as context, task criticality, and user differences (e.g., expertise, personality, cognitive abilities, and transient states like confusion or cognitive load). Our vision is that of a personalized XAI, endowing AI agents with the ability to understand when, and how to provide explanations to their end users.

As a step toward this vision, in this paper, we present and evaluate an explanation functionality for the hints provided in the Adaptive CSP (ACSP) applet, an Intelligent Tutoring System (ITS) that helps students learn an algorithm to solve constraint satisfaction problems. ITS research investigates how to create educational systems that can model students' relevant needs, states, and abilities (e.g., domain knowledge, meta-cognitive abilities, affective states) and how to provide personalized instruction accordingly [10]. We chose to focus on an ITS in this paper because—despite increasing interest in XAI research encompassing applications such as recommender systems [4], [11]–[14], office assistants [8], and intelligent everyday interactive systems (i.e. Google Suggest, iTunes Genius, etc.) [7]—thus far there has been limited work on XAI for ITS. Yet, an ITS's aim of delivering highly individualized pedagogical interventions makes the educational context a high-stake one for AI, because such interventions may have a potentially long-lasting impact on people's learning and development. If explanations can increase ITS's transparency and interpretability, this might improve both their pedagogical effectiveness as well as the acceptance from both students and educators [3].

---

[1] There is also extensive research on XAI for AI practitioners and model developers (e.g. [1], [2]) but in this paper we focus on end users of AI models and applications,

Related research has looked at the effects of having an ITS show its assessment of students' relevant abilities via an Open Learner Model (OLM [15]), with initial results showing that this can help improve student learning (e.g., [16]) and learning abilities (e.g., ability to self-assess; [17]). There is also anecdotal evidence that an OLM can impact students' trust [18].

In this paper, we go beyond OLM and investigate the effect of having an ITS generate more explicit explanations of both its assessment of the students as well as the pedagogical actions that the ITS puts forward based on this assessment. We also look at whether there is an impact of specific user characteristics on explanation usage and effectiveness.

A formal comparison of students interacting with the ACSP applet with and without the explanation functionality shows that the explanations available improve students' *trust* in the ACPS hints, *perceived usefulness of the hints*, and *intention to use* the system again. Our results also show the impact of user characteristics on how much students look at explanations when they are available, as well as on their learning gains, which constitutes useful insights to inform the design of personalized XAI for ITS.

Despite the fact that varied reactions to explanations have been observed with several AI-driven interactive systems (e.g., [4], [5], [7], [9]), thus far there has been limited work looking at linking these reactions to user characteristics. Existing results on explanations in recommender systems have shown an impact of Need for Cognition [12] (a personality trait [19]), of Openness (also personality traits [20]) and music sophistication [21], and user decision-making style (rational vs. intuitive) [22] on explanations in recommender systems; Naveed et at. [23] of shown impact of perceived user expertise for explanations of an intelligent assistant. Our results contribute to this line of research by: (i) looking at explanations for a different type of intelligent system (an ITS); (ii) confirming an impact of Need for Cognition; (iii) showing the effect of an additional personality trait (Conscientiousness) as well as of a cognitive ability related to reading proficiency. Thus, our findings broaden the understanding of which user differences should be further investigated when designing personalized XAI in a variety of application domains.

The rest of the paper is structured as follows. Section 2 discusses related work. Section 3 describes the ACSP and the AI mechanisms that drive its adaptive hints. Section 4 illustrates the explanation functionality we added to the ACSP and Section 5 the study to evaluate it along with the impact of user characteristics. Section 6 presents results related to usage and perception of the explanation functionality, whereas Section 7 reports results on the impact of explanations on student learning and perception of the ACSP hints. Section 8 provides a summary discussion of the results, and Section 9 wraps up with conclusions, limitations and future work.

## 2. Related Work

There are encouraging results on the helpfulness of explanations in intelligent user interfaces. For example, Kulesza et al. [5] investigated explaining the predictions of an agent that helps its users organize their emails. They showed that explanations helped participants understand the system's underlying mechanism, enabling them to provide feedback to improve the agent's predictions. Coppers et al. [24] added explanations to an intelligent translation system, to describe how a suggested translation was assembled from different sources, and showed that these explanations helped translators identify better quality translations. There are several results showing that explanations aimed at highlighting the most relevant features for given classifier predictions can increase user ability to identify the best model (e.g. [25]), and to understand model predictions on unseen instances (e.g.,[26], [27]). Recent results also show that incorporating user feedback into explanations facilitates the understanding of black box classifiers [28]. Other positive results on explanations were found in the field of recommender systems (RS, e.g., [11], [13]). In particular, Kulesza et al. [11] investigated the soundness ("nothing but the truth") and completeness ("the whole truth") of explanations in a music RS and found that explanations with these attributes helped users to build a better mental model of the music recommender.

There is, however, also research showing that explanations might not always be useful or wanted. Herlocker et al. [4] evaluated an explanation interface for an RS for movies. Although 86% of the users liked having the explanations the remaining 14% did not. Similarly, Bunt et al. [8] added explanations to a mixed-initiative system suggesting personalized interface customization, and showed 60% of users appreciated the explanations whereas others considered the explanation as common sense or unnecessary. In [7], the authors conducted a survey study asking participants if they would like to receive explanations on the workings of everyday AI-driven applications (e.g., Google Suggest, iTunes Genius), qualified as low-cost in terms of their impact on the users' stakes. Users were also asked their intuition on how the underlying AI worked. Most users had reasonable mental models of this, without the help of explanations. Only a few wanted additional information. Wang et al. [29] investigated POMDP-based algorithms for explaining a robot's decision making to a human teammate, and found that that the robot explanations could improve task performance, and foster trust but only if they facilitate decision-making vs. leaving participants ambiguous about how to act on the recommendations. The authors also found that the effects of explanations yielded only for versions of the robots with lower ability.

Some research looking at the role of user characteristics in XAI has focused on user preferences, mainly in the context of RSs. For instance, Cotter et al. [30] showed that users prefer explanations for why a recommender works the way it does vs. explanations that describe how it works, when receiving recommendations in the Facebook news feed. Kleinerman et al. [31] show that in contexts where acceptance of recommendations involves a significant emotional cost (i.e., in dating applications), users prefer reciprocal explanations which reflect not only the user's preferences, but also the other agent's preferences. Kouki et al. [32] report a crowd-sourced study showing that users prefer item-centric to user-centric or socio-centric explanations, although preference for the latter type is modulated by levels of the Neuroticism personality trait. Furthermore, users preferred textual explanations. Tsai and Brusilovsky [33] evaluated twelve visual explanations and three text-based explanations in an RS for conference attendees. Participants reported a preference for visual explanation over text-based explanation, although it was shown that the preferred explanation type was not always the most effective.

Going beyond user preferences, Millecamp et al. [12] found moderating effects of Need for Cognition (N4C, a personality trait [19]) on user confidence in playlists they built based on recommendations with and without explanations, delivered by a music RS. In this study, users had to explicitly open the explanations if they wanted to see them. In a follow-up study where explanations were always visible when available, Millecamp et al. [21] no longer fond an effect of N4C. Instead, they found effects of: (i) the personality trait Openness [20] on user intention to use the music recommender again, and on the perceived support received in discovering novel music, (ii) the level of user's music sophistication on the perceived level of decision support provided by the recommendations. Naveed et al. [22] found an impact of user decision-making style (rational vs. intuitive) on user perception of different types of explanations when looking at mocked-up recommendations for buying a camera. Schaffer et al. [23] found that explanations of the suggestions generated by an intelligent assistant that helped play a binary decision game were only useful for users who declared low ability at the game, whereas they had no impact on users who were overconfident of their ability. Tintarev and Mastoff [34] evaluated personalized explanations for recommendations, with personalization related to changing the content of explanations to mention only attributes most interesting to the user as opposed to all relevant attributes. They found that these explanations generated better user satisfaction with the explanations but not with the recommended items, compare to non-personalized explanations.

Within ITS, there has been research on increasing transparency via Open Learner Modeling, namely tools that allow learners to access the ITS's current assessment [15]. Although there is no clear understanding of how OLM can be beneficial for interpretability and explainability of ITS, there is evidence of an effect on learning. For instance, Porayska-Pomsta and Chryssafidou [17] did a preliminary evaluation of the OLM for a job interview coaching environment, with results suggesting that the OLM helped users to improve their self-perception and interview skills. Long and Aleven [16] report on the positive effect of an OLM for an ITS designed to foster student self-assessment abilities in algebra skills. There is also anecdotal evidence that an OLM can impact students' trust [18], where interestingly students trusted an ITS with an OLM more when they could not change assessment in the student model. Barria-Pineda et al. [35], add explanations to an OLM, but the explanations are essentially textual rephrasing of the OLM assessment. Our work goes beyond OLMs by investigating more explicit explanations of an ITS underlying AI mechanism.

## 3. The ACSP Applet

### *3.1 Interactive Simulation for AC-3*

The ACSP applet is an interactive simulation that provides tools and personalized support for students to explore the workings of the Arc Consistency 3 (AC-3) algorithm for solving constraint satisfaction problems [36]. AC-3 represents a constraint satisfaction problem as a network of variable nodes and constraint arcs. The algorithm iteratively makes individual arcs consistent by removing variable domain values inconsistent with a given constraint, until it has considered all arcs and the network is consistent. Then, if there remains a variable with more than one domain value, a procedure called domain splitting is applied to that variable in order to split the CSP into disjoint cases so that AC-3 can recursively solve each case. The ACSP applet demonstrates the AC-3 algorithm dynamics through interactive visualizations on graphs using color and highlighting (see Figure 1). The applet provides several mechanisms (accessible via buttons in the toolbar at the top of the ACSP interface) for the interactive execution of the AC-3 algorithm on available problems [37], including: Fine Step: goes through AC-3 three basic steps of selecting an arc, testing it for consistency, removing domain values to make the arc consistent; Direct Arc Click: allows the user to select an arc to apply all these steps at once. Auto AC: automatically fine step on all arcs one by one. Domain Split: select a variable to split on and specify a subset of its values for further application of AC-3 (see the pop-up box on the left side of Figure 1). Backtrack: recover alternative networks during domain splitting. Reset: return the graph to its initial status.

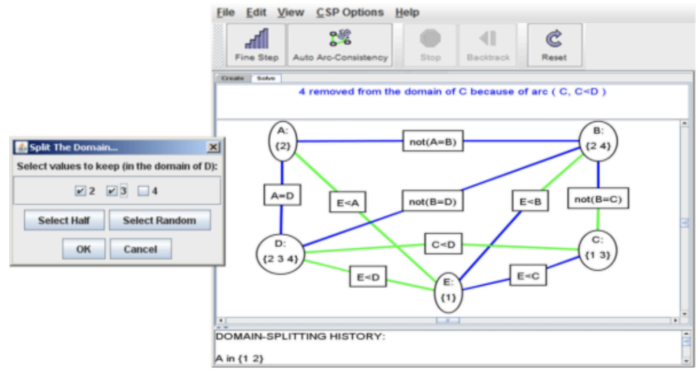

**Figure 1. The ACSP applet with an example CSP problem**

The ACSP also includes a user model that monitors how a student uses the available tools and recognizes interaction patterns that are not conducive to learning. It then leverages the predictions of the user model to generate hints guiding the student toward a more effective usage of the available tools. This user model and hint delivery mechanisms are derived based on a general framework for modeling and supporting exploratory, open-ended interactions (FUMA, Framework for User Modeling and Adaptation [38], [39]). The next two sections summarize these mechanisms since they are the targets of the explanations that we added to the ACSP.

### 3.2  Modeling User Behaviors in the ACSP

Figure 2 illustrates how the FUMA framework is integrated into the ACSP. In FUMA, the process of building a user model consists of two phases: Behavior Discovery (Figure 2 – top) and User Classification (Figure 2 – bottom right). In the following, numbers in curly braces correspond to the graph's elements in Figure 2.

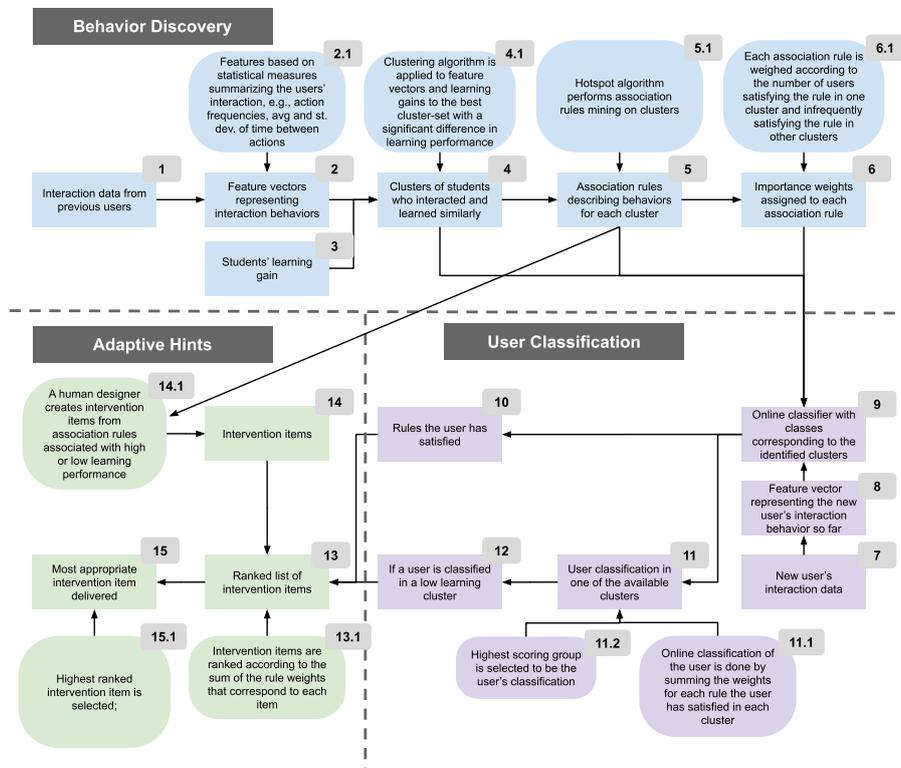

**Figure 2. ACSP User Modeling Framework broken down into three phases: Behavior Discovery, User Classification, and Adaptive Hints; rectangular nodes represent inputs and states, oval nodes represent processes**

The Behavior Discovery phase leverages existing datasets of students working with the CSP applet without adaptive hints. Data from existing interaction logs {1} is preprocessed into feature vectors consisting of statistical measures that summarize users' actions (i.e., action frequencies, time interval between actions) {2, 2.1}. Each vector summarizes the behaviors of one user. These vectors, along with data on each student's learning gains with the system[2] {3}, are fed into a clustering algorithm. The algorithm groups the feature vectors according to their similarities while also ensuring groups have significantly different learning performance. Therefore, the algorithm identifies clusters of users who interact and learn similarly with the interface {4, 4.1}. Next, association rule mining is applied to each cluster to extract its identifying interaction behaviors, i.e., rules in form of X➔ *c*, where X is a set of feature-value pairs and *c* is the predicted cluster for the data points to which X applies {5, 5.1}.

---

[2] Kardan and Conati [39] show that combining information of both interaction behavior and learning gains achieves better modeling accuracy for the subsequent User Classification phase than classifying students based on learning gains alone. Furthermore, meaningful clusters to predict learning can also be discovered when learning gains are not available [40], [41].

The rules are weighted based on how well they discriminate between the two clusters, namely based on a combination of their confidence (i.e., the relative frequency of a rule in this cluster compared to others) and their support (i.e., how frequently a rule appears in a cluster) {6, 6.1}. Based on these rules, a human designer then defines a set of hints {14, 14.1} aimed at discouraging behaviors associated with lower learning and promote behaviors associated with higher learning.

**Table 1. A subset of representative rules for HLG and LLG clusters.**

| Rules for HLG cluster |
| --- |
| Rule 1: Infrequently auto solving the CSP |
| Rule 2: Infrequently auto solving the CSP and infrequently stepping through the problem |
| Rule 3: Pausing for reflection after clicking CSP arcs |
| **Rules for LLG cluster** |
| Rule 4: Frequently backtracking through the CSP and not pausing for reflection after clicking CSP arcs |
| Rule 8: Frequently auto solving the CSP and infrequently clicking on CSP arcs |
| Rule 10: Frequently resetting the CSP |

This behavior discovery mechanism was applied to a data set of 110 users working with the CSP applet without adaptive support [34], [35]. Learning gains for these users were derived from tests on the AC-3 algorithm taken before and after using the system. From this data set, Behavior Discovery generated two clusters of users that achieved significantly different levels of learning, labeled as Higher Learning Gain (HLG) and Lower Learning Gain (LLG). A total of four and fifteen rules were found for the HLG and LLG, respectively, a selection of which is presented in Table 1. The hints that were derived from these rules are listed in Table 2.

**Table 2. Hint descriptions**

| |
|---|
| Use Direct Art Click more often; |
| Spend more time after performing Direct Arc Clicks; |
| Use Reset less frequently; |
| Use Auto Arc-consistency less frequently; |
| Use Domain Splitting less frequently; |
| Spend more time after performing Fine Steps; |
| Use Back Track less frequently; |
| Use Fine Step less frequently; |
| Spend more time after performing reset |

User Classification is the second phase involved in building the ACSP applet's user model (Figure 2 – bottom right). In this phase, the clusters, association rules, and corresponding rule weights extracted in the Behavior Discovery phase are used to build an online classifier {9}. As a new user interacts with the ACSP applet, the classifier predicts the user's learning after every action. This is done by (i) incrementally building a feature vector for the new user, based on the interface actions seen so far {7, 8} and (ii) classifying this vector in one of the available clusters using Associative Classification [42] {11, 11.1, 11.2, 12}. Specifically, cluster membership scores are calculated based on the values of the rules that are satisfied in each cluster [43]. The membership function returns a score $S_A$ for a given cluster A as follows:

$$S_A = \frac{\sum_{i=1}^{m} Test(r_i) \times W_{r_i}}{\sum_{i=1}^{m} W_{r_i}}, Test(r_i) = \begin{cases} 0, if\ r_i\ is\ satisfied \\ 1, otherwise \end{cases}$$

where $r_i$'s are the m rules selected as representatives for cluster A, $W_{r_i}$ is the corresponding rule weight, and Test($r_i$) is the function that checks the applicability of a rule to a given instance.

Note that, as the classifier predicts the user's learning after every action, the classification can change over time, depending on the evolution of the user's interaction behaviors.

In an formal evaluation of this user model, Kardan and Conati [40] found that the rule-based classification approach achieved 84.04% accuracy in classifying users into high and low learner clusters, outperforming several other supervised classification methods.

## 3.3 Adaptive Hints

In addition to classifying a user in one of the available clusters, the ACSP's user model also returns the satisfied association rules causing that classification {10}. These rules represent the characteristic interaction behaviors of a specific user so far. If the user is classified as belonging to a cluster associated with lower learning, the process of providing adaptive hints triggers (Figure 2 – bottom left). This process starts by identifying which of the hints in Table 2 should be provided when a student is classified as a lower learner at a given point of their interaction with the ACSP. More specifically, when a user is classified as a lower learner, the ACSP identifies which detrimental behaviors this user should stop performing or which beneficial behaviors they should adopt, based on the association rules that caused the classification.

Generally, a combination of rules causes the user to be classified as a lower learner, and thus, several hints might be relevant. However, to avoid confusing or overwhelming the user, the applet only delivers one hint at a time, chosen based on a ranking that reflects how predominant each of the behaviors associated with the possible hint is. After each prediction of lower learning, every item in Table 2 is assigned a score as the sum of the weights of the satisfied association rules that triggered that item and the lower learning classification {13, 13.1}. The hint with the highest score is chosen to be presented to the student {15, 15.1}.

The ACSP delivers its adaptive hints incrementally. Each hint is first delivered via a textual message that prompts or discourages a target behavior. For instance, a hint for the Use Direct Arc Click more often item in Table 2 is "Do you know that you can tell AC-3 which arc to make consistent by clicking on that arc?" (see Figure 3[3]). After receiving the hint, the student is given some time to change their behavior accordingly (a reaction window equal to 40 actions). During this time, the user model will keep updating its user classification. At the end of this time window, the user model determines whether the user has followed the hint for the target item or not, and if not, the target item is selected for delivery again, this time accompanied by stronger guidance, e.g., highlighting of relevant interface items.

---

[3] Hint appearance is slightly modified here for readability

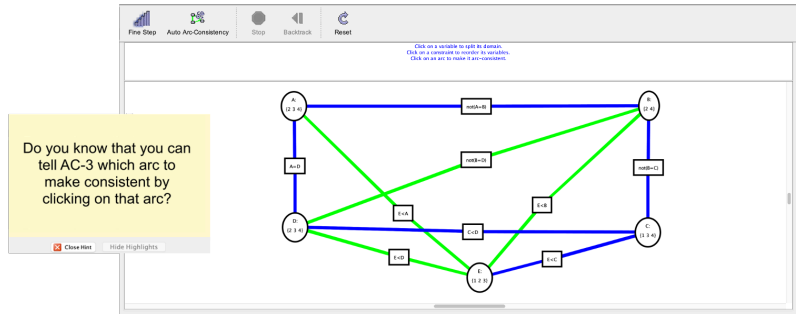

**Figure 3. The CSP applet with an example hint**

The ACSP was evaluated against a non-adaptive version with a formal study where two groups of 19 students studied three CSP problems with the adaptive and control version, respectively [39]. The study showed that students working with the ACSP learned the AC-3 algorithm better than students in the control conditions and followed on average about 73 % of the adaptive hints they received. Although these results are very positive, we want to ascertain if and how explanations of the ACSP adaptive hints might increase students' uptake and learning.

## 4. An Explanation Functionality for the ACSP applet

### 4.1 Pilot User Study

As a first step to build an explanation functionality for the ACSP applet, we wanted to gain an initial understanding of the type of explanations that students would like to have about the ACSP hints. To do so, we instrumented ACSP with a tool to elicit this information.

Namely, we added to each hint's dialogue box a button ("explain hint") that enables a panel (shown in Figure 4), allowing students to choose one or more of the following options for explanations they would have liked for these hints: (i) *why* the system gave this hint; (ii) *how* the system chose this hint; (iii) some other explanation about this hint (including a text field for user input). We also provided an option to say that no explanation for this hint was wanted. We chose to explicitly list only two types of explanations to avoid excessive complexity, and we focused on "why" and "how" explanations because they have been identified by previous work as being commonly appreciated by users (e.g. Castelli et al. [44] and Cotter et al. [30]).

**Figure 4. Interface used in the pilot study to collect feedback on which explanations users want for the ACSP hints.**

We ran a pilot study [45] with nine university students with adequate prerequisites to use the ACSP applet (as done in [39]). We told participants that we were looking for feedback on how to enrich the ACSP applet with explanations for its hints, and showed them how to provide the feedback via the "explain hint" functionality described above. During their interaction with the ACSP, the participants accessed the "explain hint" functionality for 51% of the hints delivered. Of these responses, 21 asked for a *why* explanation (48%), 14 for a *how* explanation (32%), and 6 for no explanation (14%). There were only 3 responses asking for other types of explanations: one wasn't further qualified using the free-form text box (see Figure 4), the other two in fact related to the hints, not the explanations (one asked for the hints to be more transparent, the other asked for more diverse hints). These results confirm that participants are generally interested in explanations, although to different extents, which is consistent with findings from Castelli et al. [44] and Cotter et al. [30] in terms of users preferring *why* explanations, followed by *how* explanations.

Based on the results of this pilot study, we designed and implemented an explanation functionality that conveys to the ACSP users the motivations (*why*) and processes used

(*how*) for each of the hints they receive. Essentially, these explanations should provide the ACSP users with insights on the user modeling and hint provision mechanisms described in Section 3. Although there are several other types of explanations that should be investigated (see [46]–[48]), given the results from existing literature and our pilot study, we argue that the *why* and *how* explanations we investigate in this paper are a meaningful place to start from.

## *4.2 Design Criteria*

As guidance for the explanation design, we rely on some of the criteria articulated by Kulesza et al. [5]. Specifically, in principle we want our explanations to be:

- *Iterative*, namely accessible at different levels of detail based on the user's interest
- *Sound*, namely conveying an accurate, not simplified nor distorted description of the relevant mechanisms
- *Complete*, namely exposing all aspects of the relevant mechanisms
- *Not overwhelming*, namely comprehensible and not conducive to excessive cognitive load or other negative states such as confusion and frustration

There is a trade-off that needs to be made between complying to the requirements of soundness and completeness, and avoiding that the explanations become overwhelming. The iterative criterion is an important means to achieve this trade-off, and it has a predominant role in the ACSP explanation functionality. Even so, the AI driving the ACSP hints is a complex combination of three different algorithmic components (behavior discovery, user classification, and hint selection, see Section 3), and finding a suitable way to explain these mechanisms in a manner that is accessible incrementally proved to be extremely challenging. To determine the explanation's content, the authors discussed the ACSP AI components at length and decided to start designing and evaluating a version of the explanation functionality that sacrifices completeness when it is needed to avoid excessive complexity. We do so by prioritizing *why* over *how* explanations, following the results from the pilot study described in the previous section. The rationale for this choice is to start evaluating a meaningful, albeit incomplete, set of explanations, see if and how they can impact student learning and experience with the ACSP, and eventually use these results to inform the design of future iterations of the explanation functionality.

Based on this strategy, we identified three self-contained *why* explanations, as well as three *how* explanations, described in Section 4.3. These explanations aim to help the ACSP users gain a *global understanding* of the AI driving the ACSP hints, as well as a *local* understanding of the specific hints generated. We derived these explanations from the graph in Figure 2, which represents all the inputs and states (rectangular nodes) involved in the hint computation as well as the processes (oval nodes) that generate each state from preceding ones. We use the states in Figure 2 to justify specific aspects of the rationale for hint computation (*why* explanations[4]) and the processes to explain *how* some of the relevant algorithm components work, both in general as well as instantiated to the current hint that is being explained.

We then came up with several designs to structure and navigate through these explanations, which we prototyped using a tool called Marvel to create fast wire-frame interfaces for the different designs. Test piloting the different designs revealed that the most intuitive and easy to use navigation is the tab-based design illustrated in the next section.

## 4.3 Navigation and Content

We structured the explanation interface around three tabs, each providing a self-contained, incremental part of the explanation for a given hint, as shown in the schematic flowchart in Figure 5. Each tab displays a *why* explanation; for one of these *why* explanations (tab in Figure 5 (B)), users can ask for more details on *how* three specific aspects where computed (Figure 5 (D)–(F)).

---

[4] Using Dennet's distinction between *why* explanations that address "how come" (i.e. narrate a process), vs "what for" (i.e. illustrate an objective) [49], our *why* explanations are mostly "how come" explanations, except for a "what for" explanation that opens the first explanation window: *My goal is to help you use the ACSP applet to your full potential (see section 4.3.1)*

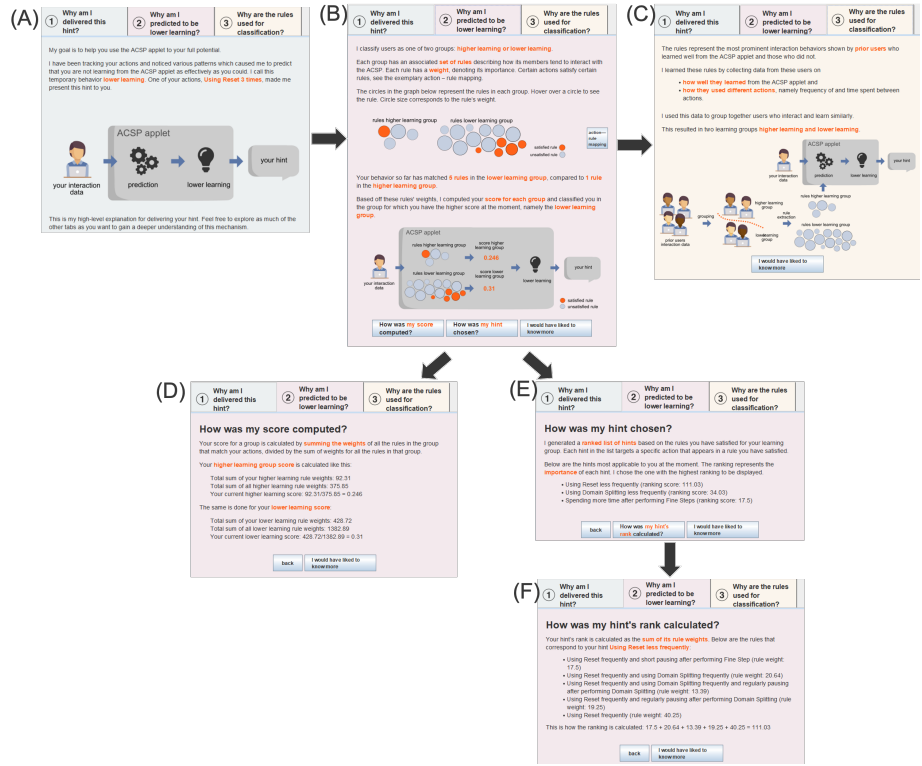

**Figure 5.** Flow Chart of Explanation Navigation (A) Why am I delivered this hint? (B) Why am I predicted to be lower learning? (C) Why are the rules used for classification? (D) How was my score computed? (E) How was my hint chosen? (F) How was my hint's rank calculated?

We refer to the different parts of the explanation as pages labelled as WhyHint, WhyLow, WhyRules, HowScore, HowHint, and HowRank in the rest of the paper. The content of each page, not readable in Figure 5, will be illustrated later in this section. Since, as we discuss in the previous section, these pages do not explain in full the ACSP's User Modeling and Behavior Discovery, users can provide feedback on the explanation's content by using a button labeled "*I would have liked to know more*" that is accessible on every page.

As mentioned above, we built these six pages of explanations from the graph in Figure 2. We selected and assembled various elements of the graph to create sound and coherent incremental explanations that the user can access at will. Note that we were originally hoping to have a one-to-one mapping between elements in the graph and explanation pages, but quickly realized that this would result in explanations that were too fragmented. Instead, we structured the explanation pages around three tabs, each providing a self-contained incremental part of the explanation for a given hint, as shown in Figure 5 (A-C). The rest of this section provides the full content of each explanation page, consisting of text and accompanying visualizations.

The explanations in the following sections are exemplary for a hint stating, "You have used the Reset button excessively. I recommend that you limit your usage of this action." The user can activate the explanation functionality once a given hint has been delivered, by clicking the button labeled "Why am I delivered this hint?"(see Figure 6).

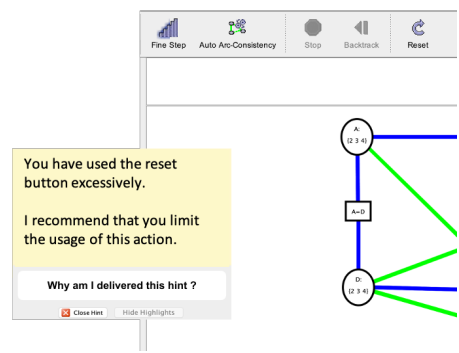

**Figure 6. The explanation functionality is triggered by clicking on the button "Why am I delivered this hint?"**

In response to request, the explanation window appears, with the first tab to the left active, as shown in Figure 5(A). The next three subsections describe the *why* explanations provided in the three tabs, as well any *how* explanation that can be requested from there. The numbers appearing in brackets in the text correspond to the graph's elements in Figure 2 that are discussed by that text. These numbers have been added here for illustration, they are not present in the explanations seen by the users. It is important to note that each explanation is dynamically derived from the current state of the underlying ACSP AI models, as they evolve based on the user's real-time interaction.

### 4.3.1 Why am I delivered this hint?

The explanation in this tab provides a high-level explanation of the user classification component and how it is linked to the hint received (see Figure 7).

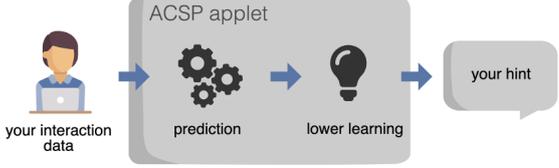

**Figure 7: Content of the "WhyHint" explanation page**

Note that, although most of this page illustrate general aspects of the rationale for hint provision (where the first sentence "My goal is…" is a *what for* type of explanation as per [49] that gives the objective of the hints), the last sentence ("One of your actions…") provides information that is specific to this user.

### *4.3.2* **Why am I predicted to be lower learning?**

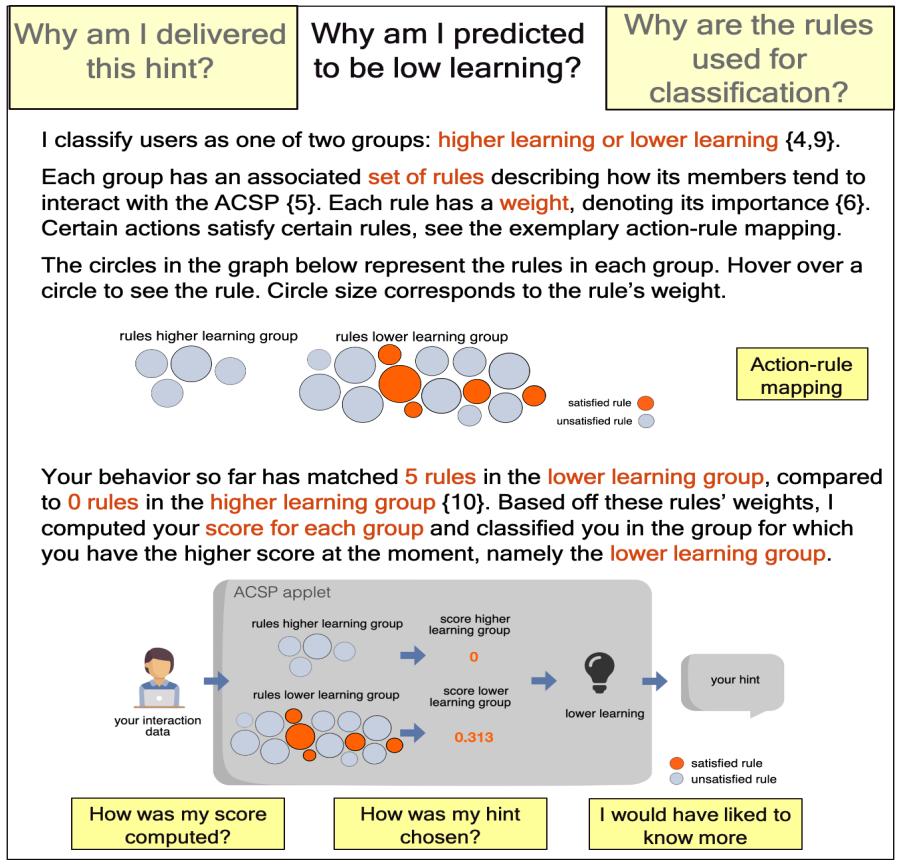

**Figure 8: Content of the "WhyLow" explanation page**

Selecting the second tab in the interface "*Why am I predicted to be lower learning?*", will give access to a more specific explanation on why the ACSP user model came up with this classification (Figure 8). Within this tab, the user can access an additional visualization linking their actions to the satisfied rules and their weights (not shown). Users can also choose to ask more details on (i) *how* their scores for each group were computed and (ii) *how* the specific hint delivered was selected (buttons at the bottom of Figure 8) resulting pages Figure 9 and Figure 10 respectively. Their content is presented below.

| Why am I delivered this hint? | Why am I predicted to be low learning? | Why are the rules used for classification? |

How was my score for each group computed?

Your score for a group is calculated by summing the weights of all the rules in the group that match your actions, divided by the sum of weights for all the rules in that group {11.1}.

Your higher learning group score is calculated like this:

Total sum of your higher learning rule weights: 0
Total sum of all higher learning rule weights: 376
Your current higher learning score: 0/376 = 0

The same is done for your lower learning score:

Total sum of your lower learning rule weights: 432
Total sum of all lower learning rule weights: 1383
Your current lower learning score: 432/1383 = .313

[back] [I would have liked to know more]

Figure 9. Content of the "HowScore" explanation page

| Why am I delivered this hint? | Why am I predicted to be low learning? | Why are the rules used for classification? |

How was my hint chosen?

I generated a ranked list of hints {13} based on the rules you have satisfied for your learning group {10}. Each hint in the list targets a specific action that appears in a rule you have satisfied.

Below are the hints most applicable to you at the moment. The ranking represents the importance of each hint. I chose the one with the highest ranking to be displayed {15, 15.1}.

- Using Reset less frequently (ranking : 98)
- Using Auto Arc Consistency less frequently (ranking:87)
- Spending more time after performing Fine Steps (ranking: 18)

[back] [How was my hint's rank calculated?] [I would have liked to know more]

Figure 10. Content of the "HowHint" explanation page

From the page in Figure 10, the user can navigate further to read more about how their hint's rank was computed, shown in Figure 11 below.

> **Why am I delivered this hint?** | **Why am I predicted to be low learning?** | **Why are the rules used for classification?**
>
> **How was my hint's rank calculated?**
>
> Your hint's rank is calculated as the sum of its rule weights {13.1}. Below are the rules that correspond to your hint Using Reset less frequently:
>
> - Using Reset less frequently and short pausing after performing Fine Step (rule weight: 18)
> - Using Auto Arc-Consistency frequently and using Reset frequently (rule weight: 21)
> - Using Reset frequently and regularly pausing after performing Domain Splitting (rule weight: 19)
> - Using Reset frequently (rule weight: 40)
>
> This is how the ranking is calculated: 18 + 21 + 29 + 40 = 109
>
> back | I would have liked to know more

**Figure 11. Content of the "HowRank" explanation page**

### 4.3.3 Why are the rules used for classification?

Selecting the third tab in the interface "*Why are the rules used for classification?*", will provide a high-level description of the Behavior Discovery phase, including background information on the data used to create the classifier and how this relates to what has already been explained (Figure 12). There is no information specific to the current user in this page.

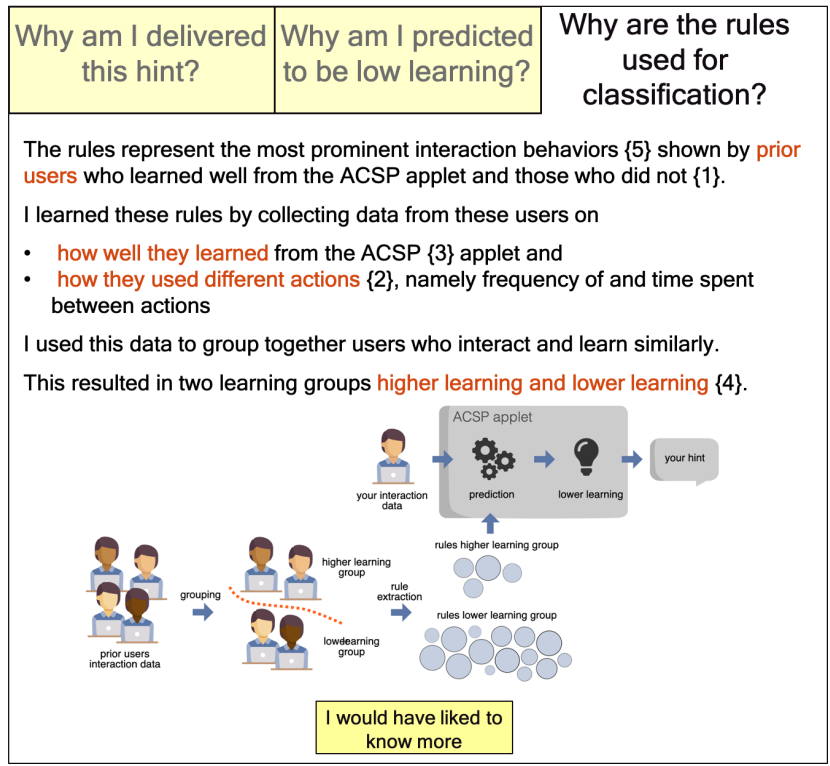

Figure 12. Content of the "WhyRules" explanation page

Note that in this tab, we could have enabled explanations on how different parts of the Behavior Discovery process work, e.g., clustering and rules extraction. However, because explaining these algorithms can be quite complicated, here is where we chose to give up explanation completeness and see how users react to this choice in the formal study described in Section 5.

## 5. User Study

This section describes the user study that we conducted to investigating whether
1) explanations on the ACSP hints influence how students perceive and learn from the adaptive hints.
2) usage and perception of the explanations depend of specific user characteristic.

The study followed a between subject design with two conditions in which participants interacted with the ACSP with and without explanation functionality (*explanation* and *control* condition, respectively).

## 5.1 Participants and Procedure

72 participants (37 female, 35 male) were recruited through advertising at our campus. They were required to have enough computer science knowledge to learn the concept of CSPs, e.g., basic graph theory and algebra, and to not have colorblindness.

The procedure for our study followed the one used in [39] to evaluate the ACSP applet hints, with minor modifications to cover the evaluation of the explanation and user characteristics. The study task was to use the ACSP applet to understand how the AC-3 algorithm solves three CSP problems [39]. Participants were told that the ACSP would provide adaptive hints during their interaction. Participants in the *explanation condition* were furthermore informed that that they could access explanations on *why* and *how* the hints were provided. Participants in this condition were shown how to access the explanation functionality but were told that it was up to them to decide whether to use it or not.

The experimental procedure was as follows: (1) participants took tests on user characteristics (see next section); (2) studied a textbook chapter on the AC-3 algorithm; (3) wrote a pre-test on the concepts covered in the chapter; (4) watched an introductory video on how to use the main functionalities of the ACSP applet; (5) used the ACSP applet to solve three CSPs; (6) took a post-test analogous to the pre-test; and (7) answered a post-questionnaire (see section 6.3). The study took between 2.5 and 3 hours in total. Participants were compensated with $30.

While the participants interacted with the ACSP applet, their gaze was tracked with a Tobii Pro X3 T-120 camera-based remote eye-tracker, embedded in the study monitor, with a sampling rate of 120 Hertz. Each participant underwent a short calibration session prior to the start of the interaction.

Of the 72 study participants 25 did not receive any hints during their interaction with the ACSP because the system assessed that they did not need help to learn effectively. Since these participants did not experience hints, and essentially worked with the ACSP applet as if there were no added AI-driven mechanisms, they are not included in the analysis presented in subsequent sections. This leaves 30 participants in the *explanation* condition and 17 participants in the *control* condition who are included in the analysis and results presented here[5].

## 5.2 User Characteristics

Study participants took a battery of well-established psychology tests to measure a set of *cognitive abilities* that may affect the processing of the explanations' textual content (i.e. *Reading Proficiency* [50]) and images (i.e. *Perceptual Speed* [51] and *Visual Working Memory* [52]). They also took test for *traits* that may affect the perception of the explanations and hints, based on existing literature. These are

- Need for Cognition [19] included because it was found to have an impact on explanation effectiveness in [12].
- The five personality dimensions of the Big 5 Model, namely Agreeableness, Conscientiousness, Extraversion, Neuroticism, and Openness [20]. These were included because at least one of them was found to have an impact on explanation preference in [32].
- Two dimensions of Curiosity [53]: *Joyous Exploration*, and *Deprivation Sensitivity*. We added these traits because some users in the pilot study in Section 4.1 mentioned curiosity when asked reasons for wanting explanations

Table 3 reports the UC investigated, with their relevant details

---

[5] The difference in size between the two experimental conditions is due to the fact that we gave priority to have a sufficient number of participants in the explanation group to perform a meaningful analysis of how the explanations are accessed and perceived, accounting for the fact that some users might not access the explanation functionality.

**Table 3: Set of user characteristics, with tests used to measure them, test score range, and statistics for the scores of the participants**

| User Char | Definition | Test | Test range | Mean ± SD |
|---|---|---|---|---|
| Reading Proficiency | Vocabulary size and reading comprehension ability in English [50] | *X_Lex Vocabulary Test* [50] | 0;100 | 80.32 ±15.4 |
| Perceptual Speed | Speed in scanning/comparing figures or symbols, or carrying out other very simple tasks involving visual perception [51] | *P-3 Identical Pictures Test* [51] | 0;72 | 47.03 ±7.86 |
| Visual Working Memory | Quantity of visual information (e.g., shapes and colors) that can be temporarily maintained or manipulated in working memory [52] | *Colored Squares Sequential Comparison Task (uncued)* [52] | 0;6 | 2.46 ±0.92 |
| Need for Cognition | Extent to which individuals are inclined towards effortful cognitive activities [19] | *Need for Cognition Scale* [19] | -36;36 | 1.26 ±4.5 |
| Agreeableness | An agreeable person is fundamentally altruistic, sympathetic to others and eager to help them [54] | *Ten-Item Personality Inventory* [20] | 1;7 | 4.39 ±0.98 |
| Conscientiousness | Refers to self-control and the active process of planning, organizing and carrying out tasks [54] | *Ten-Item Personality Inventory* [20] | 1;7 | 5.11 ±1.24 |
| Extraversion | Includes traits such as sociability, assertiveness, activity and talkativeness [54] | *Ten-Item Personality Inventory* [20] | 1;7 | 4.06 ±1.44 |
| Neuroticism | General tendency to experience negative affects such as fear, sadness, embarrassment, anger, guilt and disgust [54] | *Ten-Item Personality Inventory* [20] | 1;7 | 4.39 ±1.36 |
| Openness | Openness to Experience includes active imagination, aesthetic sensitivity, attentiveness to inner feelings, a preference for variety, intellectual curiosity and independence of judgement [54] | *Ten-Item Personality Inventory* [20] | 1;7 | 5.06 ±1.05 |
| Joyous Exploration | Extent to which one derives positive emotions from learning new information and experiences [53] | *Five-Dimensional Curiosity Scale* [53] | 0;1 | 0.72 ±0.15 |
| Deprivation Sensitivity | Desire to reduce gaps in knowledge because they generate feelings of anxiety and tension [53] | *Five-Dimensional Curiosity Scale* [53] | 0;1 | 0.64 ±0.17 |

## 5.3 Questionnaires

Given that one of the goals of this study is to understand if and how explanations change a user's perception of the ACSP AI-driven adaptive hints, all study participants filled out a questionnaire designed to measure such perception. The questionnaire included the items listed in Table 4, which were rated on a 5-point scale ranging from strongly disagree (1) to strongly agree (5).

**Table 4. Perception of Hints Questionnaire Items**

| |
|---|
| **Items on Usefulness** |
| H1. I would choose to have the hints again in the future. |
| H2. I am satisfied with the hints. |
| H3. The hints were helpful for me. |
| **Items on Intrusiveness** |
| H4. The hints distracted me from my learning task. |
| H5. The hints were confusing. |
| **Items on Hints Understanding and Trust** |
| H6. I understand why hints were delivered to me in general |
| H7. I understand why specific hints were delivered to me |
| H8. I trust the system to deliver appropriate hints |
| H9. Given my behavior, I agree with the hints that were delivered to me |

The items in Table 4 were selected from a variety of sources including the Usefulness, Satisfaction, and Ease of use (USE) questionnaire[55], as well as established XAI literature (e.g., [8], [12], [56]). The first three items target the general perceived usefulness of the hints by gauging users' intention to have them again in the future, satisfaction with, and perceived helpfulness of the hints. Items H4 and H5 gauge potential intrusiveness of the hints in terms of distraction and confusion, as done by Kardan et al. [39] in the first evaluation of the ACSP applet. The last four items measure the user understanding of why the hints were delivered, as well as user trust that the hints are appropriate.

We also collected subjective feedback on the explanation functionality (see Table 5) from participants in the *explanation* group. The first five items are similar to those for the hints questionnaire, gauging general usefulness and intrusiveness of the explanations. An additional item was included for intrusiveness (E6), namely being *overwhelming*, because avoiding overwhelming explanations is one of the specific design criteria for explanations in [5]. The last three items in the explanation questionnaire evaluate the usability of the explanation functionality in terms of clarity of accessibility, navigation, and content to ensure that none of these factors inhibited users from using the explanations.

**Table 5. Explanation Questionnaire Items**

| |
|---|
| **Items on Usefulness** |
| E1. I would choose to have the explanations again in the future. |
| E2. I am satisfied with the explanations. |
| E3. The explanations were helpful for me. |
| **Items on Intrusiveness** |
| E4. The explanations distracted me from my learning task. |
| E5. The explanations were confusing. |
| E6. I found the explanations overwhelming. |
| **Items on Usability** |
| E7. It was clear to me how to access the explanations. |
| E8. The explanation navigation was clear to me. |
| E9. The explanation content (i.e., wording, text, figures) was clear to me. |

Participants in the *explanation* group who chose not to view the explanations did not receive the questionnaire in Table 5. Instead, they answered the open-ended question "Please describe why you did not access the explanation, using the button 'Why was I delivered this hint?'".

## 6. Usage and perception of explanations

This section presents results on how the 30 participants in the explanation condition of the study used and perceived the explanation functionality. Of these 30 participants, 24 accessed the explanation functionality at least once. Section 6.1 reports results on how these users rated the explanation functionality, Section 6.2 looks at how and how much they use it, and Section 6.3 reports whether the amount of usage is modulated by the user characteristics described in Section 5.2. Section 6.4 discusses the 6 participants that never accessed explanations.

### 6.1 Subjective Ratings on the Explanations

The analysis of the questionnaire items on the explanation functionality (Table 5 in Section 5.3) reveals that users were in general positive about it. This can be seen in

Figure 13 (left), with the high ratings for the items related to intention to use and satisfaction, whereas helpfulness has more room for improvement. The low ratings for distraction, confusion, and overwhelming (Figure 13 - center) also speak in favor of the explanation functionality. Furthermore, most users strongly agreed that the explanation functionality is clear in access, navigation, and content (Figure 13 - right), suggesting a strong usability.

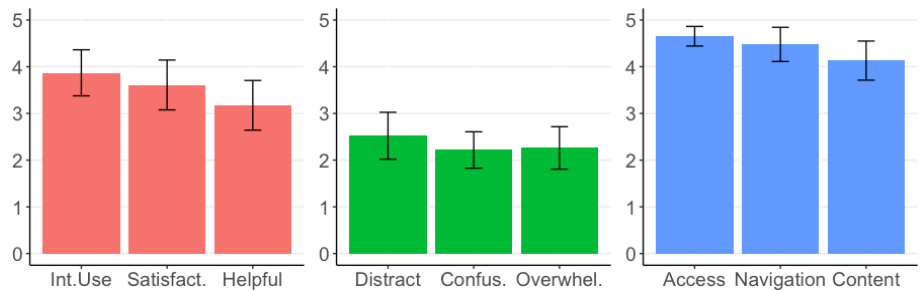

**Figure 13. Subjective ratings on the explanations. Answers to items E1-E3 (left graph), E4-E6 (center), E7-E9 (right) in Table 5**

## 6.2 Interaction with the Explanation Interface

In this section we discuss if and how the 24 participants in the explanation condition accessed the explanations. To do so, we logged the participants' interaction events and extracted a variety of explanation-related actions. These actions include:

- *Explanation initiation*: starting the explanation for a given hint.
- *Page access*: accessing any one of the explanation's pages; during each explanation initiation, there can be multiple page accesses for each available page
- *Explanation type accessed*: accessing one of the six types of explanations available (see Figure 5); thus, the number of explanation type accessed ranges from 1 to 6.

We also tracked the amount of time participants actually looked at the explanations they accessed, using the data from the eye-tracker described in Section 5.1. Eye-trackers capture the position and duration of user *fixations*, namely episodes of user gaze focusing on specific points on the screen. Our eye-tracker also allows for identifying when fixations happen in specific regions of the screen, thus, in order to capture how much time our participants spent looking at the ACSP explanation, we defined the following:

- *Attention to explanations per hints received*: total duration of fixations on any of the explanation pages that a participant opened during their interaction with the ACSP, divided by the total number of hints received.

Participants in the explanation condition received on average 2.6 hints, with large standard deviation (2.2) and range (minimum of one hint and maximum of 11). Figure 14 shows, for each participant, the number of hints received as well as their *why* and *how* page accesses.

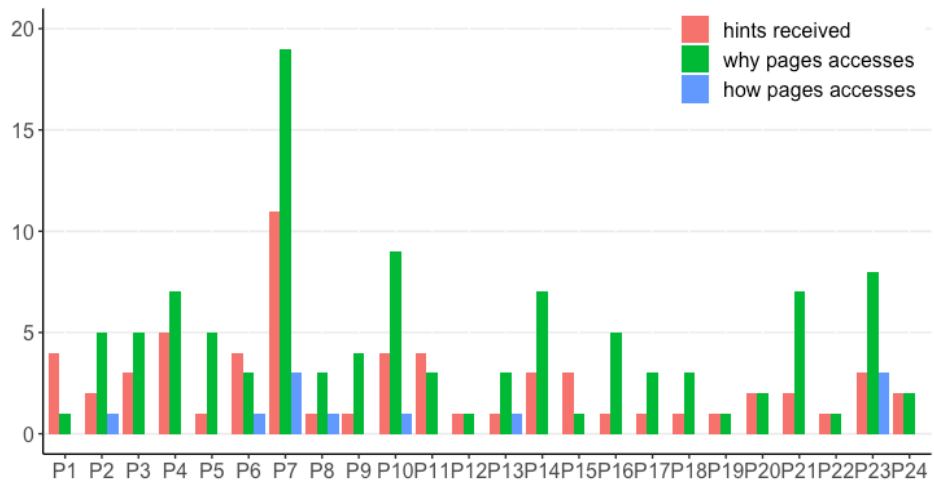

**Figure 14. Number of hints, *why* page accesses, and *how* page accesses per participant**

This figure gives an overall sense of the substantial variability with which these 24 participants engaged with the explanation functionality, whereas Table 6 provides detailed summative statistics on the usage of the explanations.

**Table 6. Summative statistics on usage of the explanation**

|  | Mean | SD | Min | Max |
|---|---|---|---|---|
| Hints before first explanation initiation | 1.08 | 0.28 | 1 | 2 |
| Explanation initiations per hints received | 0.75 | 0.29 | 0.25 | 1.00 |
| Number of page accesses per initiation | 3.11 | 1.64 | 1 | 7 |
| Attention to explanations per hint received | 38.5s | 39.18s | .65s | 140s |
| Explanation types accessed | 2.79 | 1.28 | 1 | 5 |

The first two rows in Table 6 concern how participants initiated explanations in response to hints. The first row shows that participants tended to initiate explanations for the first time as soon as they received the first hint, or the second hint at the latest. The ratio of explanation initiations per number of hints received (second row) is 0.75 on average, i.e., participants initiated the explanation for 3/4 of the hints received, indicating that some participants were eager to view explanations and went back to the explanations for subsequent hints.

The last three rows in Table 6 give a sense of how much participants actually explored the explanation interface. There were, on average, 3 page accesses per each initiation (third row), with a minimum of 1 and maximum of 7 accesses. Participants spent an average of 38.5s per hint received looking at the explanation interface (fourth row), with a notable standard deviation of 39.18s and a range between less than a second and above 2 minutes. Finally, of the 6 different types of explanation pages available, close to 3 were accessed on average, with a range between 1 and 5.

Figure 15 visualizes the proportion of each explanation type accessed.

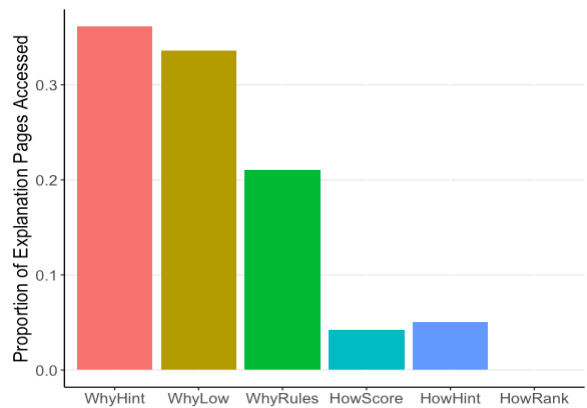

**Figure 15. Proportion of number of accesses for each type of explanation page**

This gives the picture of users mainly accessing the first two *why* pages, as taking together the proportions for WhyHint and WhyLow (Figure 7 and Figure 8) makeup about two thirds of total accesses. The third type of *why* explanation (WhyRules, Figure 12) takes up most of remaining third. As far as the *how* explanations are concerned, the proportions of accesses are similar for HowScore and HowHint (Figure 9 and Figure 10), the two types that can be directly accessed from WhyLow, and reach zero for the page that explains how a hint's rank was computed (HowRank, Figure 11). These patterns of access are consistent with the results of the pilot study in Section 4.1 related to users being generally more interested in *why* than *how* explanations, but they also show the influence of the position of each page in the navigation structure, with access decreasing for pages further down the access paths.

Only one participant made use of the "I would have liked to know more" button, wishing for more details on the rules.

## 6.3 Effects of User Characteristics on Attention to the Explanation Interface

To investigate whether user characteristics (UC from now on) influence participants tendency to make use of the ACSP explanations, we look at the attention measures defined in the previous section, namely the number of page accesses per initiation, and the attention to explanations per hint received, or *Explanation Attention* from now on. We chose these two dependent measures as representative of the amount of effort a participant was willing to put into exploring the explanation interface. For each of these dependent measures, we run a group of ANCOVAs by selecting each of the UC (see Section 5.2) as a covariate, one at a time. We run separate models for each covariate to avoid overfitting our models by including all co-variates at once. Since there was no strong correlation[6] among the tested UCs, each model can be considered as an independent analysis on the impact of the target individual difference on the dependent

---

[6] All pairwise Pearson correlation coefficients were small or medium (r<.5, following [57])

measure[7]. We report significant results at the level of p < .05 after correcting for multiple comparisons using Benjamini and Hochberg [58] adjustment to control for the family-wise false discovery rate (FDR) for the 2 dependent measures tested.

We found that *Need for Cognition* (*N4C* from now on) has a statistically significant effect on *Explanation Attention* with a medium effect size (F(1,22)=8.51, p=.008, $\eta_p^2$=.28)[8]. Figure 16 shows the directionality of the effect where, for visualization purposes only, we binarized the users into low and high values of N4C by using a median split on their N4C test scores.

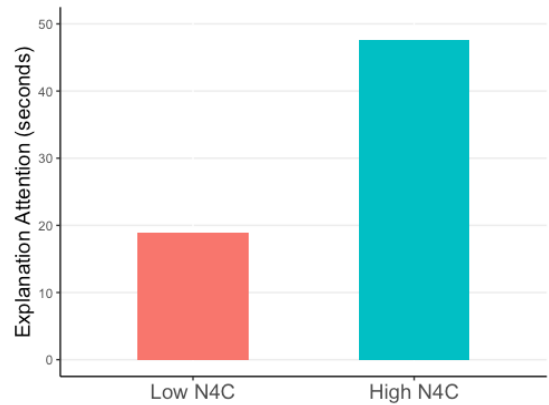

**Figure 16. Effect of N4C on Explanation Attention**

Participants with a high level of N4C show higher values of Explanation Attention than participants with low N4C. This finding suggests that high N4C users are more prone to leverage the ACSP expatiations, which aligns with the group's characteristics to be inclined toward effortful cognitive activities (see Section 5.2). The fact that low N4C users tend to pay less attention to the explanation might be detrimental for them if the explanations turn out to be useful for improving user learning and/or satisfaction with the ACSP applet, as we investigate in Section 7.

---

[7] We will use this approach for all further analyses that look at the impact of user characteristics in subsequent sections

[8] Following [59], [60] we consider effect sizes to be high for *d, r, $\eta_p^2$* >.5; medium for *r* >.3 and *d, $\eta_p^2$* >.2; small otherwise

*6.4 Users who did not access the explanations*

The six users who did not access the explanation functionality received between 1 and 3 hints (M=1.5, sd=.83). Instead of answering the questionnaire on explanations (Table 5), they answered an open ended question asking for the reasons of not viewing any explanations. Their responses (summarized in Table 7) indicate that explanations were not accessed either because the hints were clear enough, or because they were not wanted in the first place. Given the low numbers of these users, we cannot perform any formal analysis on the possible influence of user characteristics on their unwillingness to ask for explanations, but this is indeed an interesting question for future investigations.

**Table 7. Reasons for not accessing explanations**

| **Hints are self-explanatory** |
|---|
| "The text in the body of the hint conveyed enough information to me, so there was no need to access the explanation of the hint" |
| "The hint was self-explanatory, so I didn't consider looking into the details in the explanation" |
| "The hint is very straightforward and I was expecting it to pop up" |
| **Hints are not wanted in the first place** |
| "I don't need the hint to finish the task" |
| "I prefer to just skip and continue my job" |
| "I did not find the hint useful to start with so I thought the explanation would be the same" |

## 7. Effects of Explanations on Learning and Users' Perception of the Hints

In this section, we investigate whether leveraging the ACSP explanations has an effect on student learning and on their perception of the ACSP hints, and whether these effects may be affected by user characteristics.

We do so by comparing learning and hints perception between participants in the explanation condition who accessed the explanation functionality, and the participants in the control conditions. We exclude from the analysis the six users in the explanation condition who did not access any explanation (Section 6.4) because explanations could not have any effect on their learning and hint perception. This leaves 24 users in the *explanation* condition (those discussed in sections 6.1-6.3) to be compared against the 17 users in the *control* condition.

## 7.1 Effect of Explanations on Learning

As a dependent measure for this analysis we use *percentage learning gains (PLG*, a standard measure to gauge learning in ITS*)*: the ratio of the gain between the post-test and pre-test a participant took in the study, over the maximum possible gain. A two sample unpaired t-test, between the PLGs of the control and explanation group showed no statistically significant difference in the scores for participants in the control (M=.45) and the explanation (M=.49) conditions (t(25.7) = .36, p=.7, d=.12). Next, we evaluate whether the effect of explanations on learning gains may be modulated by user characteristics (UC), by checking for interaction effects between UCs and condition on PLGs. For each of the UCs in Section 5.2, we ran an ANCOVAs with that UC as a covariate, condition (explanation vs control) as factor and PLGs as the dependent variable.

We found a significant interaction effect for Conscientiousness (F(1,37)=4.76, p=.035, $\eta_p^2$=.11), which is illustrated in Figure 17 by binarizing users into high and low test scores for Conscientiousness, based on a median split.

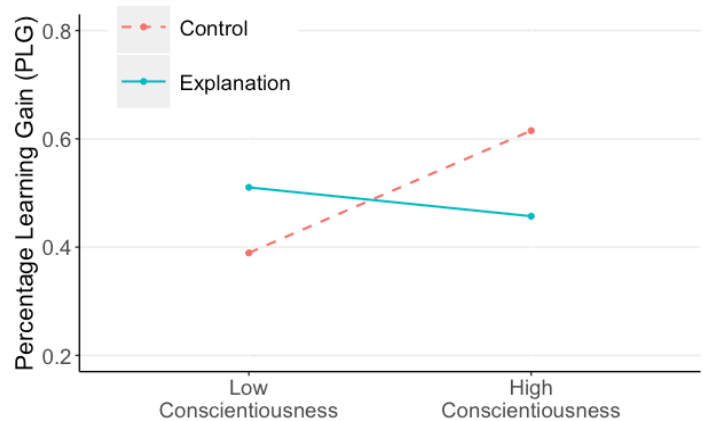

**Figure 17. Interaction effect between level of user Conscientiousness (low vs. high) and condition (explanation vs. control) on PLG.**

The figure shows a trend of users with *low* conscientiousness learning better with explanations than without, whereas the opposite is true for users with *high* conscientiousness. Conscientiousness is a personality trait related to being diligent, perform tasks well, and being efficient and organized toward achieving one's goals [54]. Thus, it is possible that, for users with higher levels of contentiousness, receiving hints from the ACSP on how to make the interaction with the system more effective is sufficient to make them comply, whereas users with lower contentiousness need more motivation to do so, and this might be provided by the explanations.

This is of course just a hypothesis, that needs to be further investigated. However, the interaction effect that we found for conscientiousness is interesting in itself because it supports the idea that XAI functionalities should be personalized. Here, for instance, if the ACSP knows that a user has low conscientiousness, it may proactively propose explanations because of their positive effect we found on learning for low conscientiousness users. On the contrary, for high conscientiousness users, explanations might even be withdrawn since they have a possible detrimental effect on their learning.

## 7.2 Effect on Explanations on Perception of the ACSP Hints

To evaluate whether the presence of explanations has a main effect on how users perceive the ACSP, and specifically its AI-driven hints, for each subjective measure collected in the questionnaire on hints (see Table 4 in Section 5.3), we compare the ratings from users in the explanation and control conditions using Wilcoxon-Mann-Whitney test, suitable for analysis of ordinal ratings. Figure 18 shows the average ratings of control vs explanation condition for each of the nine subjective measures. We report significant results at the level of $p < .05$ after correcting for multiple comparisons using Benjamini and Hochberg [58] adjustment for the 9 subjective measures tested. We found a significant effect of experimental condition on three of the subjective measures. H1, H3, and H8: (marked by *** in Figure 18). Namely, compared to users in the control condition, users in the explanation condition reported:

- higher **intention to use** the hints (H1) ($z=2.85$, $p=.0044$, $r=.44$).
- higher ratings for the hints being **helpful** to them (H3) ($z=3.16$, $p=.0016$, $r=.49$).
- higher **trust** (H8) in the system to deliver appropriate hints ($z=2.62$, $p=.0088$, $r=.41$).

Although there was no significant effect of explanations on the remaining six items of the hint questionnaire, the trend is generally in favor of having vs. not having the explanations, as shown in Figure 18.

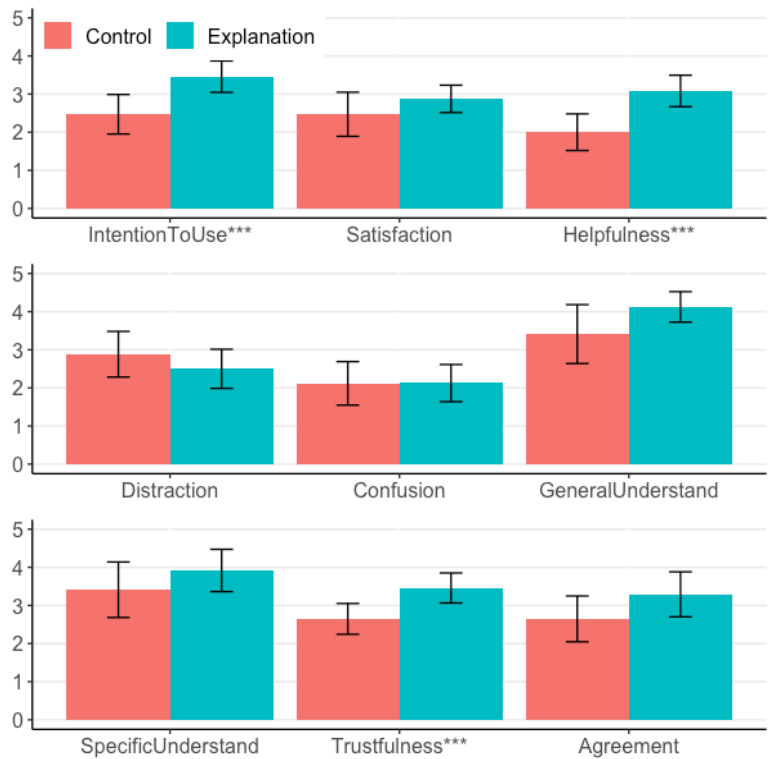

**Figure 18. Subjective ratings in the control and explanation conditions on the perception of hints. From top to bottom: answers to items H1-H3, H4-H6, H7-H9 in Table 4. Statistical difference between the two conditions is indicated with\*\*\***

Next, as we did for learning gains, we ascertain whether the users' perception of the hints with and without explanations is modulated by the user characteristics tested in the study (see Section 5.2), by checking for interaction effects between UC and conditions on the 9 hint perception measures. To do so, for each hint perception measure we ran a proportional odds regression for ordinal (Likert-scale) variables, with that measure as dependent variable, condition (explanation vs. control) as the independent variable, and each of the UCs as covariate, one at a time. We report significant interaction effects between user characteristic and condition on a given dependent variable at p<.05 after correcting for multiple comparisons using Benjamini and Hochberg [58] adjustment for 9 dependent measures.

We found a significant interaction effect between *reading proficiency* and condition on **confusion** (H5: $\chi^2$=9.97, *p*=.0016, r=.49). The directionality of the effect is illustrated in Figure 19, where users are divided into high and low reading proficiency using a median split on their test scores for this measure. As the figure shows, users with high reading proficiency found the hints less confusing when provided with explanations as compared to no explanations, whereas the trend is reversed for low reading proficiency users. This effect makes sense, given that the explanations contain substantial amounts of text. Users with higher levels of reading proficiency are better able to process this text, thus the explanations help them resolve confusion they may have with the hints, but make hints more confusing for those users less skilled at reading. This effect provides additional evidence that personalization is important in XAI. Here, it suggests that we need to investigate alternative ways for the ACSP to present the explanations to low reading proficiency users (e.g., less text, more graphics).

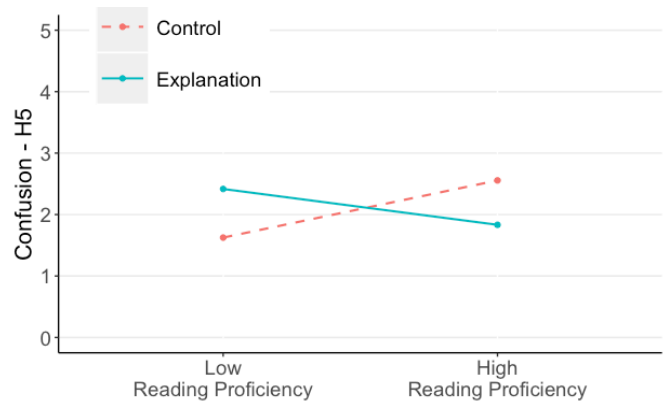

**Figure 19. Interaction between level of user reading proficiency (low vs. high) and condition (explanation vs. control) on the ratings provided in H5.**

## 8. Discussion

Designing an explanation functionality that conveys to ACSP users at least some of the AI mechanisms driving the ACSP adaptive hints is challenging, because of the complexity of such mechanisms. The study presented in this paper aimed to ascertain if our first attempt at an explanation functionality that illustrates such AI mechanisms can have a positive impact on how students perceived the ACSP hints and learn from them.

Our results indicate that accessing these explanations lead to students reporting higher trust in the ACSP pedagogical hints, as well as feeling more helped by them. Furthermore, users were more willing to want the ACSP hints in the future when they accessed explanations about the mechanism driving its hints.

We also provided evidence that the ACSP explanation functionality might be more effective if it is personalized to specific user needs.

First, our results show that users with lower values of Need for Cognition (N4C) tend to pay less attention to the explanations than their high N4C counterparts. Since our study showed that the ACSP explanations were beneficial to improve some aspects of user perception of the ACSP hints, and these hints have been shown to promote learning [39], it is worthwhile exploring ways to make the ACSP encourage low N4C users to pay more attention to its explanations, as a way to increase hints' acceptance by these users. Interestingly, N4C was also shown to be relevant for personalized explanations in a very different domain, namely recommender systems for music [12], suggestig that this is a user trait worthy of further investigation for personlized XAI across domains.

Second, we found that whether the ACSP explanations improve student learning depend to some extent on a student's level of conscientiousness, where students with low conscientiousness benefit from having the explanations whereas their high-level counterparts don't. Similarly, we found that whether the ACSP explanations help clarify the ACSP hints (i.e., reduce student confusion about the hints) depends on student level of reading proficiency, where students with high levels of this measure benefit from having the explanations, whereas their low levels counterpart don't, likely due to the large amount of text currently used in the explanations. These findings suggest modifying the ACSP explanation functionality so that it proactively encourages users who are known to have low contentiousness or high reading proficiency to access explanations. They also uncover the need to investigate how to reduce the possibly negative effects of explanations on users with high contentiousness and low reading proficiency, for instance by discouraging explanation access or by understanding how to make them useful for these users.

Information on the relevant user characteristics can be collected upfront using the standard tests we used in the study, after which personalization could be enabled by setting related parameters in the ACSP. However, existing research has shown that user characteristics such as some personality traits [61] and cognitive abilities [62] can be detected in real-time from interaction data, which could be leveraged as part of the user modeling performed as students work with the ACSP applet.

It should be noted that there might be further effects of explanations and user characteristics that our study did not have enough power to detect. Our sample sizes were limited by a number of factors: participants had to have specific prerequisites to be able to interact meaningfully with the ACSP; the length of the study (over two hours), which was a deterrent for participation even if we offered monetary compensation; the use of eye-tracking, which limited running participants in parallel; having to discard participants who did not receive hints from the ACSP. These factors are a necessary consequence of investigating explanations with an AI system that tackles a challenging task (supporting human leaning), evaluating them in a realistic context of usage, and collecting rich data beyond simple interaction events. We see our results as providing important initial insights of what explanations can do in this context.

## 9. Conclusions, Limitations and Future Work

This paper contributes to understanding the need for personalization in XAI. Although the importance of having AI artifacts that can explain their actions to their users is undisputed, there is mounting evidence that one-size-fits-all explanations are not ideal, and that explanations may need to be tailored to several factors including context, task criticality, and specific user needs. Our research focuses on the latter and, in this paper, we present a case study that investigates the need for personalization for XAI in Intelligent Tutoring Systems (ITS). Although there is booming interest in AI for Education and there has been research on how to increase ITS transparency via Open Learner Models, thus far work on enabling ITS to provide explicit explanations on the AI underlying their user modeling and decision making has been preliminary at best.

The contributions of this paper include:

- An explanation functionality enabling incremental access to *why* and *how* explanations for the adaptive hints generated by the ACSP, an ITS that supports learning via an interactive simulation.

- Results showing positive effect of the explanation functionality on user's perception of the ACSP adaptive hints.

- Results showing how some user characteristics modulate the effect of explanations on student's learning and perception of the ACSP adaptive hints, providing insights on how personalization might bolster explanation effectiveness.

Our results are the first to show that explanations of an ITS pedagogical actions can have a positive effect. Furthermore, our findings indicate that user characteristics modulate the effect of explanations which, in line with previous work [12], [32], calls for the need of personalized XAI.

This research is of course just one step toward understanding the value of XAI in ITS, and the need for personalization. There are several limitations in our work that should be addressed in future iterations. First, our results show that users accessed explanations related to why a hint was generated substantially more than explanations about how it was selected. These results, however, are likely affected by how accessible each explanation page was in our explanation functionality; thus we plan to investigate designs that will provide more conclusive results on this point. Second, our explanations are quite complex and verbose. Although this reflects the complexity of the underlying AI mechanism, and these explanations still generated positive results, it is worthwhile exploring how to make them more lightweight and engaging, possibly by including more visuals and animations. Third, there might be other effects of explanations and user characteristics that our study did not have enough power to detect. Fourth, we only looked at *why* and *how* explanations, but there are several other types that should be investigated. We tried to elicit from users which other explanations they would like, but we asked via open-ended textual questions that did not generated much feedback, thus we will explore other ways to identify explanation types worth adding Finally, ACSP is designed for students with specific computer science background, who might be more interested in understanding the underlying AI via explanations than other users with different backgrounds. It is crucial to investigate explanations in ITS designed to work with less technology-savvy students, as they might generate very different reactions than the ones we observed.

Additional future work includes performing further analysis of the eye-tracking data we collected, which could unveil the specific attention allocation strategies used to access the different components of the explanation functionality, and the role that user characteristics play in these strategies. Next, we plan to add to the ACSP the capability to personalize its explanations to the different levels of user characteristics identified as relevant in this paper, as well as investigate if these user characteristics can be automatically predicted from interaction [61] or eye tracking data [62], [63]. Finally, we want to look at how short-term user states such as confusion, cognitive load, and affective reactions might be taken into account for personalizing explanations.

## Acknowledgements

This work was supported by the Natural Sciences and Engineering Research Council of Canada NSERC [Grant #22R01881]

# References

[1] A. Barredo Arrieta *et al.*, "Explainable Artificial Intelligence (XAI): Concepts, taxonomies, opportunities and challenges toward responsible AI," *Information Fusion*, vol. 58, pp. 82–115, Jun. 2020, doi: 10.1016/j.inffus.2019.12.012.

[2] F. Hohman, M. Kahng, R. Pienta, and D. H. Chau, "Visual Analytics in Deep Learning: An Interrogative Survey for the Next Frontiers," *IEEE Transactions on Visualization and Computer Graphics*, vol. 25, no. 8, pp. 2674–2693, Aug. 2019, doi: 10.1109/TVCG.2018.2843369.

[3] C. Conati, K. Porayska-Pomsta, and M. Mavrikis, "AI in Education needs interpretable machine learning: Lessons from Open Learner Modelling," *ICML Workshop on Human Interpretability in Machine Learning*, Jun. 2018, Accessed: Mar. 22, 2020. [Online]. Available: http://arxiv.org/abs/1807.00154.

[4] J. L. Herlocker, J. A. Konstan, and J. Riedl, "Explaining collaborative filtering recommendations," in *Proceedings of the 2000 ACM conference on Computer supported cooperative work*, Philadelphia, Pennsylvania, USA, Dec. 2000, pp. 241–250, doi: 10.1145/358916.358995.

[5] T. Kulesza, M. Burnett, W.-K. Wong, and S. Stumpf, "Principles of Explanatory Debugging to Personalize Interactive Machine Learning," in *Proceedings of the 20th International Conference on Intelligent User Interfaces*, Atlanta, Georgia, USA, Mar. 2015, pp. 126–137, doi: 10.1145/2678025.2701399.

[6] S. Kraus *et al.*, "AI for Explaining Decisions in Multi-Agent Environments," *Thirty-Fourth AAAI Conference on Artificial Intelligence*, 2020.

[7] A. Bunt, M. Lount, and C. Lauzon, "Are explanations always important? a study of deployed, low-cost intelligent interactive systems," in *Proceedings of the 2012 ACM international conference on Intelligent User Interfaces*, Lisbon, Portugal, Feb. 2012, pp. 169–178, doi: 10.1145/2166966.2166996.

[8] A. Bunt, J. McGrenere, and C. Conati, "Understanding the Utility of Rationale in a Mixed-Initiative System for GUI Customization," in *User Modeling 2007*, Berlin, Heidelberg, 2007, pp. 147–156, doi: 10.1007/978-3-540-73078-1_18.

[9] K. Ehrlich, S. E. Kirk, J. Patterson, J. C. Rasmussen, S. I. Ross, and D. M. Gruen, "Taking advice from intelligent systems: the double-edged sword of explanations," in *Proceedings of the 16th international conference on Intelligent user interfaces*, Palo Alto, CA, USA, Feb. 2011, pp. 125–134, doi: 10.1145/1943403.1943424.

[10] B. P. Woolf, *Building Intelligent Interactive Tutors: Student-centered Strategies for Revolutionizing E-learning*. Morgan Kaufmann, 2010.

[11] T. Kulesza, S. Stumpf, M. Burnett, S. Yang, I. Kwan, and W.-K. Wong, "Too much, too little, or just right? Ways explanations impact end users' mental models," in *2013 IEEE Symposium on Visual Languages and Human Centric Computing*, Sep. 2013, pp. 3–10, doi: 10.1109/VLHCC.2013.6645235.

[12] M. Millecamp, N. N. Htun, C. Conati, and K. Verbert, "To explain or not to explain: the effects of personal characteristics when explaining music recommendations," in *Proceedings of the 24th International Conference on Intelligent User Interfaces*, Marina del Ray, California, Mar. 2019, pp. 397–407, doi: 10.1145/3301275.3302313.

[13] I. Nunes and D. Jannach, "A systematic review and taxonomy of explanations in decision support and recommender systems," *User Model User-Adap Inter*, vol. 27, no. 3, pp. 393–444, Dec. 2017, doi: 10.1007/s11257-017-9195-0.

[14] M. Wiebe, D. Y. Geiskkovitch, and A. Bunt, "Exploring User Attitudes Towards Different Approaches to Command Recommendation in Feature-Rich Software," in *Proceedings of the 21st International Conference on Intelligent User Interfaces*, Sonoma, California, USA, Mar. 2016, pp. 43–47, doi: 10.1145/2856767.2856814.

[15] S. Bull and J. Kay, "SMILI☺: a Framework for Interfaces to Learning Data in Open Learner Models, Learning Analytics and Related Fields," *Int J Artif Intell Educ*, vol. 26, no. 1, pp. 293–331, Mar. 2016, doi: 10.1007/s40593-015-0090-8.

[16] Y. Long and V. Aleven, "Enhancing learning outcomes through self-regulated learning support with an Open Learner Model," *User Model User-Adap Inter*, vol. 27, no. 1, pp. 55–88, Mar. 2017, doi: 10.1007/s11257-016-9186-6.

[17] K. Porayska-Pomsta and E. Chryssafidou, "Adolescents' Self-regulation During Job Interviews Through an AI Coaching Environment," in *Artificial Intelligence in Education*, Cham, 2018, pp. 281–285, doi: 10.1007/978-3-319-93846-2_52.

[18] A. Mabbott and S. Bull, "Student Preferences for Editing, Persuading, and Negotiating the Open Learner Model," in *Intelligent Tutoring Systems*, Berlin, Heidelberg, 2006, pp. 481–490, doi: 10.1007/11774303_48.

[19] J. T. Cacioppo, R. E. Petty, and C. F. Kao, "The Efficient Assessment of Need for Cognition," *Journal of Personality Assessment*, vol. 48, no. 3, pp. 306–307, Jun. 1984, doi: 10.1207/s15327752jpa4803_13.

[20] S. D. Gosling, P. J. Rentfrow, and W. B. Swann, "A very brief measure of the Big-Five personality domains," *Journal of Research in Personality*, vol. 37, no. 6, pp. 504–528, Dec. 2003, doi: 10.1016/S0092-6566(03)00046-1.

[21] M. Millecamp, N. N. Htun, C. Conati, and K. Verbert, "What's in a User? Towards Personalising Transparency For Music Recommender Interfaces," in *Proceeding of the 28th conference on User Modeling, Adaptation and Personalization*, Jul. 2020, Accessed: Apr. 22, 2020. [Online]. Available: https://lirias.kuleuven.be/3010826.

[22] S. Naveed, T. Donkers, and J. Ziegler, "Argumentation-Based Explanations in Recommender Systems: Conceptual Framework and Empirical Results," in *Adjunct Publication of the 26th Conference on User Modeling, Adaptation and Personalization*, Singapore, Singapore, Jul. 2018, pp. 293–298, doi: 10.1145/3213586.3225240.

[23] J. Schaffer, J. O'Donovan, J. Michaelis, A. Raglin, and T. Höllerer, "I can do better than your AI: expertise and explanations," in *Proceedings of the 24th International Conference on Intelligent User Interfaces*, Marina del Ray, California, Mar. 2019, pp. 240–251, doi: 10.1145/3301275.3302308.

[24] S. Coppers *et al.*, "Intellingo: An Intelligible Translation Environment," in *Proceedings of the 2018 CHI Conference on Human Factors in Computing Systems*, Montreal QC, Canada, Apr. 2018, pp. 1–13, doi: 10.1145/3173574.3174098.

[25] M. T. Ribeiro, S. Singh, and C. Guestrin, "'Why Should I Trust You?': Explaining the Predictions of Any Classifier," in *Proceedings of the 22nd ACM SIGKDD International Conference on Knowledge Discovery and Data Mining*,

San Francisco, California, USA, Aug. 2016, pp. 1135–1144, doi: 10.1145/2939672.2939778.
[26] M. T. Ribeiro, S. Singh, and C. Guestrin, "Anchors: High-Precision Model-Agnostic Explanations," in *Thirty-Second AAAI Conference on Artificial Intelligence*, Apr. 2018, Accessed: Apr. 22, 2020. [Online]. Available: https://www.aaai.org/ocs/index.php/AAAI/AAAI18/paper/view/16982.
[27] A. R. Akula, S. Wang, and S.-C. Zhu, "CoCoX: Generating Conceptual and Counterfactual Explanations via Fault-Lines," in *Thirty-Fourth AAAI Conference on Artificial Intelligence*, Feb. 2020, Accessed: Apr. 22, 2020. [Online]. Available:
https://www.aaai.org/ocs/index.php/AAAI/AAAI18/paper/view/16982.
[28] H. Lakkaraju, E. Kamar, R. Caruana, and J. Leskovec, "Faithful and Customizable Explanations of Black Box Models," in *Proceedings of the 2019 AAAI/ACM Conference on AI, Ethics, and Society*, Honolulu, HI, USA, Jan. 2019, pp. 131–138, doi: 10.1145/3306618.3314229.
[29] N. Wang, D. V. Pynadath, and S. G. Hill, "The Impact of POMDP-Generated Explanations on Trust and Performance in Human-Robot Teams," in *Proceedings of the 2016 International Conference on Autonomous Agents & Multiagent Systems*, Singapore, Singapore, May 2016, pp. 997–1005, Accessed: Apr. 22, 2020. [Online].
[30] K. Cotter, J. Cho, and E. Rader, "Explaining the News Feed Algorithm: An Analysis of the 'News Feed FYI' Blog," in *Proceedings of the 2017 CHI Conference Extended Abstracts on Human Factors in Computing Systems*, Denver, Colorado, USA, May 2017, pp. 1553–1560, doi: 10.1145/3027063.3053114.
[31] A. Kleinerman, A. Rosenfeld, and S. Kraus, "Providing explanations for recommendations in reciprocal environments," in *Proceedings of the 12th ACM Conference on Recommender Systems*, Vancouver, British Columbia, Canada, Sep. 2018, pp. 22–30, doi: 10.1145/3240323.3240362.
[32] P. Kouki, J. Schaffer, J. Pujara, J. O'Donovan, and L. Getoor, "Personalized explanations for hybrid recommender systems," in *Proceedings of the 24th International Conference on Intelligent User Interfaces*, Marina del Ray, California, Mar. 2019, pp. 379–390, doi: 10.1145/3301275.3302306.
[33] C.-H. Tsai and P. Brusilovsky, "Evaluating Visual Explanations for Similarity-Based Recommendations: User Perception and Performance," in *Proceedings of the 27th ACM Conference on User Modeling, Adaptation and Personalization*, Larnaca, Cyprus, Jun. 2019, pp. 22–30, doi: 10.1145/3320435.3320465.
[34] N. Tintarev and J. Masthoff, "Evaluating the effectiveness of explanations for recommender systems," *User Model User-Adap Inter*, vol. 22, no. 4, pp. 399–439, Oct. 2012, doi: 10.1007/s11257-011-9117-5.
[35] J. Barria-Pineda, K. Akhuseyinoglu, and P. Brusilovsky, "Explaining Need-based Educational Recommendations Using Interactive Open Learner Models," in *Adjunct Publication of the 27th Conference on User Modeling, Adaptation and Personalization*, Larnaca, Cyprus, Jun. 2019, pp. 273–277, doi: 10.1145/3314183.3323463.
[36] D. L. Poole and A. K. Mackworth, *Artificial Intelligence: Foundations of Computational Agents*. Cambridge University Press, 2010.

[37] S. Amershi, G. Carenini, C. Conati, A. K. Mackworth, and D. Poole, "Pedagogy and usability in interactive algorithm visualizations: Designing and evaluating CIspace," *Interacting with Computers*, vol. 20, no. 1, pp. 64–96, Jan. 2008, doi: 10.1016/j.intcom.2007.08.003.

[38] S. Kardan, "A data mining approach for adding adaptive interventions to exploratory learning environments," University of British Columbia, 2017.

[39] S. Kardan and C. Conati, "Providing Adaptive Support in an Interactive Simulation for Learning: An Experimental Evaluation," in *Proceedings of the 33rd Annual ACM Conference on Human Factors in Computing Systems*, Seoul, Republic of Korea, Apr. 2015, pp. 3671–3680, doi: 10.1145/2702123.2702424.

[40] S. Kardan and C. Conati, "Comparing and Combining Eye Gaze and Interface Actions for Determining User Learning with an Interactive Simulation," in *User Modeling, Adaptation, and Personalization*, Berlin, Heidelberg, 2013, pp. 215–227, doi: 10.1007/978-3-642-38844-6_18.

[41] S. Lallé and C. Conati, "A Data-Driven Student Model to Provide Adaptive Support during Video Watching Across MOOCs," in *Artificial Intelligence in Education*, Ifrane, 2020, p. to appear.

[42] F. Thabtah, "A review of associative classification mining," *The Knowledge Engineering Review*, vol. 22, no. 1, pp. 37–65, Mar. 2007, doi: 10.1017/S0269888907001026.

[43] X. Yin and J. Han, "CPAR: Classification based on Predictive Association Rules," in *Proceedings of the 2003 SIAM International Conference on Data Mining*, 0 vols., Society for Industrial and Applied Mathematics, 2003, pp. 331–335.

[44] N. Castelli, C. Ogonowski, T. Jakobi, M. Stein, G. Stevens, and V. Wulf, "What Happened in my Home? An End-User Development Approach for Smart Home Data Visualization," in *Proceedings of the 2017 CHI Conference on Human Factors in Computing Systems*, Denver, Colorado, USA, May 2017, pp. 853–866, doi: 10.1145/3025453.3025485.

[45] V. Putnam and C. Conati, "Exploring the Need for Explainable Artificial Intelligence (XAI) in Intelligent Tutoring Systems (ITS)," in *IUI Workshops*, 2019.

[46] D. Wang, Q. Yang, A. Abdul, and B. Y. Lim, "Designing Theory-Driven User-Centric Explainable AI," in *Proceedings of the 2019 CHI Conference on Human Factors in Computing Systems*, Glasgow, Scotland Uk, May 2019, pp. 1–15, doi: 10.1145/3290605.3300831.

[47] T. Miller, "Explanation in artificial intelligence: Insights from the social sciences," *Artificial Intelligence*, vol. 267, pp. 1–38, Feb. 2019, doi: 10.1016/j.artint.2018.07.007.

[48] B. Y. Lim, A. K. Dey, and D. Avrahami, "Why and why not explanations improve the intelligibility of context-aware intelligent systems," in *Proceedings of the SIGCHI Conference on Human Factors in Computing Systems*, Boston, MA, USA, Apr. 2009, pp. 2119–2128, doi: 10.1145/1518701.1519023.

[49] D. C. Dennett, *From Bacteria to Bach and Back: The Evolution of Minds*. W. W. Norton & Company, 2017.

[50] P. Meara, *EFL vocabulary tests*. ERIC Clearinghouse New York, 1992.

[51] R. B. Ekstrom, D. Dermen, and H. H. Harman, *Manual for kit of factor-referenced cognitive tests*, vol. 102. Educational testing service Princeton, NJ, 1976.

[52] E. K. Vogel, G. F. Woodman, and S. J. Luck, "Storage of features, conjunctions, and objects in visual working memory," *Journal of Experimental Psychology: Human Perception and Performance*, vol. 27, no. 1, pp. 92–114, 2001, doi: 10.1037/0096-1523.27.1.92.

[53] T. B. Kashdan *et al.*, "The five-dimensional curiosity scale: Capturing the bandwidth of curiosity and identifying four unique subgroups of curious people," *Journal of Research in Personality*, vol. 73, pp. 130–149, Apr. 2018, doi: 10.1016/j.jrp.2017.11.011.

[54] S. Rothmann and E. P. Coetzer, "The big five personality dimensions and job performance," *SA Journal of Industrial Psychology*, vol. 29, no. 1, pp. 68–74, Jan. 2003.

[55] A. M. Lund, "Measuring usability with the use questionnaire," *Usability interface*, vol. 8, no. 2, pp. 3–6, 2001.

[56] R. Kocielnik, S. Amershi, and P. N. Bennett, "Will You Accept an Imperfect AI? Exploring Designs for Adjusting End-user Expectations of AI Systems," in *Proceedings of the 2019 CHI Conference on Human Factors in Computing Systems*, Glasgow, Scotland Uk, May 2019, pp. 1–14, doi: 10.1145/3290605.3300641.

[57] A. Field and G. Hole, *How to Design and Report Experiments*. SAGE Publications, 2003.

[58] Y. Benjamini and Y. Hochberg, "Controlling the False Discovery Rate: A Practical and Powerful Approach to Multiple Testing," *Journal of the Royal Statistical Society: Series B (Methodological)*, vol. 57, no. 1, pp. 289–300, 1995, doi: 10.1111/j.2517-6161.1995.tb02031.x.

[59] J. Cohen, "Eta-squared and partial eta-squared in fixed factor ANOVA designs," *Educational and psychological measurement*, vol. 33, no. 1, pp. 107–112, 1973.

[60] J. Cohen, *Statistical power analysis for the behavioral sciences, Rev. ed.* Hillsdale, NJ, US: Lawrence Erlbaum Associates, Inc, 1977, pp. xv, 474.

[61] L. Küster, C. Trahms, and J.-N. Voigt-Antons, "Predicting personality traits from touchscreen based interactions," in *2018 Tenth International Conference on Quality of Multimedia Experience (QoMEX)*, May 2018, pp. 1–6, doi: 10.1109/QoMEX.2018.8463375.

[62] C. Conati, S. Lallé, Md. A. Rahman, and D. Toker, "Further results on predicting cognitive abilities for adaptive visualizations," in *Proceedings of the 26th International Joint Conference on Artificial Intelligence*, Melbourne, Australia, Aug. 2017, pp. 1568–1574, Accessed: Apr. 14, 2020. [Online].

[63] S. Lallé, C. Conati, and G. Carenini, "Predicting confusion in information visualization from eye tracking and interaction data," in *Proceedings of the Twenty-Fifth International Joint Conference on Artificial Intelligence*, New York, New York, USA, Jul. 2016, pp. 2529–2535, Accessed: Apr. 14, 2020. [Online].